%% file: neurips_2025.tex
\documentclass{article}

\usepackage[preprint]{neurips_2025}

\usepackage[utf8]{inputenc} 
\usepackage[T1]{fontenc}    
\usepackage{hyperref}       
\usepackage{url}            
\usepackage{booktabs}       
\usepackage{amsfonts}       
\usepackage{nicefrac}       
\usepackage{microtype}      
\usepackage{xcolor}         
\usepackage{bm}
\usepackage{amsmath}
\usepackage{graphicx}
\usepackage{wrapfig}        
\usepackage{siunitx}        
\usepackage{booktabs}       
\usepackage{array}

\title{ViewCraft3D: High-Fidelity and View-Consistent 3D Vector Graphics Synthesis}

\author{
    Chuang Wang \textsuperscript{1}\footnotemark[1] \;
    Haitao Zhou \textsuperscript{1}\footnotemark[1] \;
    Ling Luo \textsuperscript{2}\footnotemark[2] \;
    \textbf{Qian Yu} \textsuperscript{1}\footnotemark[2] \; \\
    {\textsuperscript{1}Beihang University \quad
    \textsuperscript{2}Ningbo University \quad
    }   \\
    \footnotemark[1]\, Equal contribution \quad
    \footnotemark[2]\, Corresponding author
}

\begin{document}

\maketitle

\begin{figure*}[h]
\centering
\vspace{-1cm}
\includegraphics[width=1\linewidth]{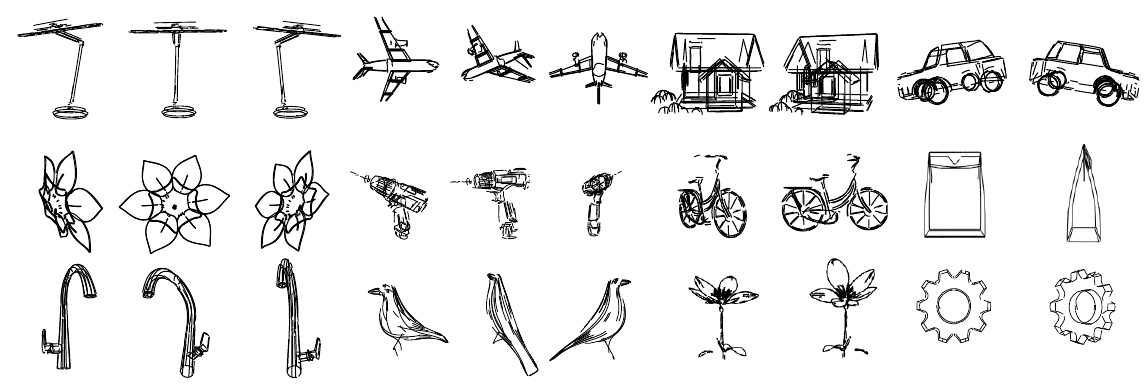}
\caption{
We propose \textit{ViewCraft3D (VC3D)}, a method to generate 3D vector graphics from a single image.
VC3D can leverage 3D prior knowledge to generate high-quality and view-consistent 3D vector graphics.
}
\label{fig:tears}
\end{figure*}

\begin{abstract}

3D vector graphics play a crucial role in various applications including 3D shape retrieval, conceptual design, and virtual reality interactions due to their ability to capture essential structural information with minimal representation.
While recent approaches have shown promise in generating 3D vector graphics, they often suffer from lengthy processing times and struggle to maintain view consistency.
To address these limitations, we propose \textbf{V}iew\textbf{C}raft\textbf{3D} (VC3D), an efficient method that leverages 3D priors to generate 3D vector graphics.
Specifically, our approach begins with 3D object analysis, employs a geometric extraction algorithm to fit 3D vector graphics to the underlying structure, and applies view-consistent refinement process to enhance visual quality.
Our comprehensive experiments demonstrate that VC3D outperforms previous methods in both qualitative and quantitative evaluations, while significantly reducing computational overhead. The resulting 3D sketches maintain view consistency and effectively capture the essential characteristics of the original objects.
Project page: \url{https://zhtjtcz.github.io/VC3D\_page/}.
\end{abstract}

\input{intro}

\input{related}

\input{method}

\input{exp}

\input{con}


\bibliography{neurips_2025}
\bibliographystyle{plain}

\end{document}

%% file: intro.tex
\section{Introduction}
\label{intro}
Three-dimensional vector graphics offer a unique balance between abstraction and comprehensibility, using minimal line elements to convey complex spatial information.
These economical representations have become integral to diverse computing applications, from improving immersive experiences in virtual environments to facilitating 3D shape retrieval and reconstruction tasks \cite{yu2021cassie, luo2020towards, luo20233d, lee2025recovering, yu2022piecewise}.
In virtual reality creation environments, 3D vector graphics serve as intuitive building blocks that allow artists to materialize spatial concepts directly within immersive spaces \cite{yu20243d, arora2021mid, yu2023videodoodles}, bridging the gap between imagination and digital realization.
Recent interactive sketching tools \cite{bae2008ilovesketch, arora2021mid, yu2021scaffoldsketch} have enhanced these creative capabilities by enabling direct manipulation in 3D space. Despite these advances, creating effective 3D vector graphics remains prohibitively difficult for non-specialists due to the intricate combination of spatial reasoning, technical interface skills, and artistic judgment required.
This expertise barrier significantly limits widespread adoption and accessibility, highlighting the need for automated approaches that can generate high-quality 3D vector graphics without requiring users to have specialized training or artistic expertise.

\begin{wrapfigure}{r}{0.48\linewidth}
    \centering
    \includegraphics[width=\linewidth]{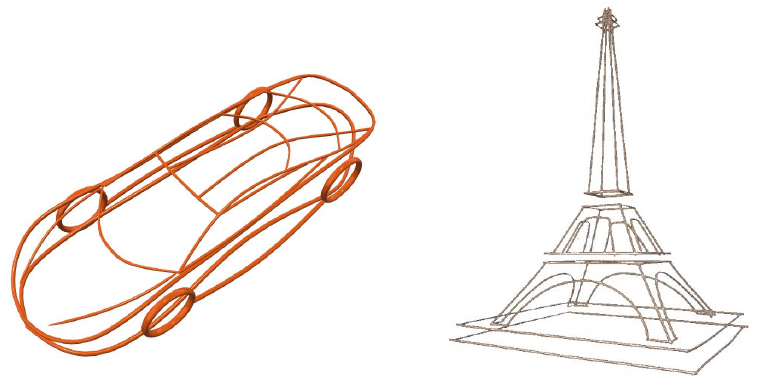}
    \caption{ Examples of VR sketches~\cite{Sketchfab}. }
    \label{fig:intro}
\end{wrapfigure}

Recent years have witnessed remarkable progress in 2D vector graphics generation.
Works like CLIPasso~\cite{vinker2022clipasso} and CLIPDraw~\cite{frans2022clipdraw} pioneered the use of CLIP's visual-semantic understanding to guide vector graphics optimization.
Building on these foundations, methods such as VectorFusion~\cite{jain2023vectorfusion}, DiffSketcher~\cite{DiffSketcher}, and SVGDreamer~\cite{svgdreamer} further leveraged diffusion models to achieve higher fidelity and controllability in vector graphics generation.
Concurrently, the field of 3D content creation~\cite{yang2024hunyuan3d, zhao2025hunyuan3d, li2025triposg, xiang2024structured, wu2024unique3d, ye2025hi3dgen} has been revolutionized by neural rendering techniques and generative models, making high-quality 3D asset creation increasingly accessible.
The convergence of these advancements has catalyzed research in 3D vector graphics, with pioneering works like 3Doodle~\cite{choi20243doodle} and Diff3DS~\cite{zhang2024diff3ds} demonstrating the feasibility of generating expressive 3D line drawings.
These approaches have achieved impressive results in creating 3D vector graphics.
However, existing methods predominantly rely on 2D generative priors—leveraging models like CLIP~\cite{radford2021learning} and diffusion model~\cite{rombach2022high} as supervision signals—while employing Score Distillation Sampling (SDS)~\cite{dreamfusion} for optimization in 2D projection space rather than directly in 3D.
These indirect approaches inherit a fundamental limitation of 2D SDS optimization: cross-view inconsistency, which constrains the ability of methods~\cite{dreamfusion, zhang2024diff3ds}—where the same 3D element appears inconsistently from different viewpoints.
Even with the use of more powerful pretrained models, these approaches often struggle to generate coherent 3D vector graphics that remain consistent across arbitrary viewpoints. For example, Diff3DS~\cite{zhang2024diff3ds} employs MVDream~\cite{shi2023mvdream} to tackle this issue, but the improvement is only partial. On the other hand, 2D priors from pretrained image generation models offer only conceptual-level guidance, lacking precise recovery of critical lines typically found in human-drawn 3D sketches, as illustrated in Figure~\ref{fig:intro}.
As a result, the generated outputs often suffer from messy strokes, missing details, and low structural fidelity.

To overcome these challenges, we propose \textbf{V}iew\textbf{C}raft\textbf{3D} (VC3D), a novel approach that leverages 3D priors for generating high-fidelity and view-consistent 3D vector graphics.
Instead of relying on optimization using 2D priors~\cite{zhang2024diff3ds,choi20243doodle}, our method is grounded in 3D geometric attributes within the 3D domain. This allows it to naturally inherit the cross-view consistency of the 3D object while faithfully preserving its spatial structure and geometric details, as illustrated in Figure~\ref{fig:tears}. Specifically, we start by reconstructing a 3D mesh using a pre-trained image-to-3D model. 
Based on the resulting mesh, we identify salient regions in 3D space that capture the object's key structural features. We then perform point-level clustering using spatial proximity and orientation alignment. 
These clusters are subsequently fitted with 3D Bézier curves, and Chamfer Distance loss is used to ensure accurate geometric approximation.
To further refine these vector graphics, we introduce a 3D score distillation sampling loss based on pretrained 3D generative models, which optimizes the Bézier curve parameters to enhance both visual quality and structural fidelity. This approach maintains view consistency by construction, as the optimization occurs directly in 3D space guided by 3D priors.

In summary, our contributions are threefold: 
\begin{itemize}
    \setlength{\itemsep}{3pt}
    \setlength{\parskip}{3pt}
    \setlength{\parsep}{3pt}
    \item We propose ViewCraft3D (VC3D), a novel framework for generating high-fidelity 3D vector graphics that leverages 3D priors rather than 2D projections;
    \item We develop a two-phase optimization approach combining geometric fitting with 3D-prior guided refinement, significantly improving visual quality.
    \item We conduct extensive experiments demonstrating that our approach outperforms existing methods in both view consistency and generation speed. The results suggest promising directions for future studies.
\end{itemize}

%% file: related.tex
\section{Related Work}
\label{related}

\subsection{2D Vector Graphics Generation}

Early approaches in 2D SVG generation, such as CLIPasso~\cite{vinker2022clipasso} and CLIPDraw~\cite{frans2022clipdraw}, utilized the visual-semantic understanding of the CLIP model~\cite{radford2021learning} to guide vector optimization.
Subsequent research introduced more sophisticated approaches, notably those employing diffusion models~\cite{ho2020denoising, rombach2022high, esser2024scaling}.
Work like VectorFusion~\cite{jain2023vectorfusion}, DiffSketcher~\cite{DiffSketcher}, SVGDreamer~\cite{svgdreamer},
and SVGDreamer++~\cite{svgdreamer++} demonstrated significant improvements in generation quality by employing Score Distillation Sampling~\cite{dreamfusion}.
This technique effectively transfers the generative capabilities of pixel-based models to the vector domain.
In addition, SVGFusion~\cite{xing2024svgfusion} explored the use of the DiT architecture~\cite{peebles2023scalable} to generate SVG.
Furthermore, specialized approaches have been developed for specific applications. These include Word-as-image~\cite{iluz2023word} for typographic design, CLIPascene~\cite{vinker2023clipascene} for scene sketching with varying abstraction levels, and VectorPainter~\cite{hu2024vectorpainter} for stylized graphics synthesis.

More recently, efforts have focused on mitigating the computational cost associated with iterative optimization.
Works based on autoregressive models, such as Iconshop~\cite{wu2023iconshop}, have demonstrated the potential for rapid generation, significantly reducing processing times.
Concurrently, the adaptation of large language models (LLMs) for SVG generation has emerged as another promising research avenue, with works like LLM4SVG~\cite{xing2024empowering} and Chat2SVG~\cite{wu2024chat2svg}.
And OmniSVG~\cite{yang2025omnisvg} attempts to employ Vision-Language Models (VLMs) as end-to-end multimodal SVG generators. Together, these recent advancements aim to ensure high-quality generation while paving the way for future extensions into 3D representations.

\subsection{Recent Advances in 3D Content Generation}

Recent years have witnessed remarkable progress in 3D content generation driven by diffusion-based approaches. Early works like Zero-123~\cite{liu2023zero} pioneered single-image view synthesis using geometric priors from diffusion models, while One-2-3-45~\cite{liu2023one} extended this to generate full 360-degree textured meshes. Multi-view consistency became a focus with MVDream~\cite{shi2023mvdream}, which serves as an implicit 3D prior through multi-view image generation, and Wonder3D~\cite{long2024wonder3d}, which employs cross-domain attention for consistent normal and color generation. Recent innovations have further elevated capabilities: Unique3D~\cite{wu2024unique3d} improved fidelity through multi-level upscaling, HunYuan3D~\cite{yang2024hunyuan3d, zhao2025hunyuan3d} achieved photorealistic quality, TripoSG~\cite{li2025triposg} utilized triplane optimization with large-scale data, and Hi3DGen~\cite{ye2025hi3dgen} enhanced geometric fidelity through normal bridging.
These cutting-edge approaches primarily focus on generating complete 3D assets with textures and materials, while our work emphasizes the creation of 3D vector graphics that maintain characteristic abstractions and representational efficiency. By leveraging the 3D understanding embedded in these advanced models, particularly TripoSG's structural representations, we guide our vector optimization toward semantically meaningful and view-consistent results.

\subsection{3D Vector Graphics Generation}

Building upon both the 2D vector graphics techniques and recent 3D generation advances discussed above, 3D vector graphics generation has emerged as a promising research direction. These representations extend the fundamental advantages of 2D vector graphics while leveraging 3D generative capabilities to model complex spatial structures and depth information. This integration enhances their utility in diverse fields, including web development and digital art. In artistic contexts, works like DreamWire~\cite{qu2024wired} and Fabricable 3D Wire Art~\cite{tojo2024fabricable} have showcased the potential of 3D vector graphics to create compelling, view-dependent visual effects, where the perceived objects change based on viewing angle.

To harness these benefits and enable such advanced applications, the development of robust 3D vector graphics generation techniques has become a key research focus.
Initial explorations in this area include 3Doodle~\cite{choi20243doodle}, which pioneered a method for generating 3D vector graphics from multi-view images of the target object.
Subsequently, Diff3DS~\cite{zhang2024diff3ds} utilized the Score Distillation Sampling  to produce 3D vector graphics conditioned on text or image input.
Dream3DVG~\cite{Dream3DVG} leverages the optimization process of a 3D Gaussian Splatting~\cite{kerbl20233d} to establish a coarse-to-fine generation approach.
However, a notable aspect of these current generative approaches is their predominant reliance on 2D view-specific loss for optimization. View-specific loss is computed independently per camera view, and gradients are aggregated across views during optimization. Without strong 3D regularization (e.g., geometry priors or multi-view constraints), this results in locally optimal solutions per view that can conflict globally.
While yielding impressive outcomes, this strategy may not fully exploit 3D spatial cues, potentially leading to challenges such as view inconsistency in the final 3D vector representations.






%% file: method.tex
\section{Methodology}
\label{method}

\subsection{Overview}

In this section, we introduce \textbf{V}iew\textbf{C}raft\textbf{3D} (\textbf{VC3D}), an optimization based method that creates a 3D vector graphic $\mathcal{S}^{3D}$ based on an input image $\bm{I}$.
We define a 3D vector graphic $\mathcal{S}^{3D}$ as a set of 3D Bézier curves $\{ C_{i} \}_{i=1}^{n}$.
The curves are defined by a set of control points $\{ P_{i,j} \}_{j=1}^{m}$, where $P_{i,j} \in \mathbb{R}^{3}$ is the $j$-th control point of the $i$-th curve.

\begin{figure}[htb]
  \centering
  \includegraphics[width=1\textwidth]{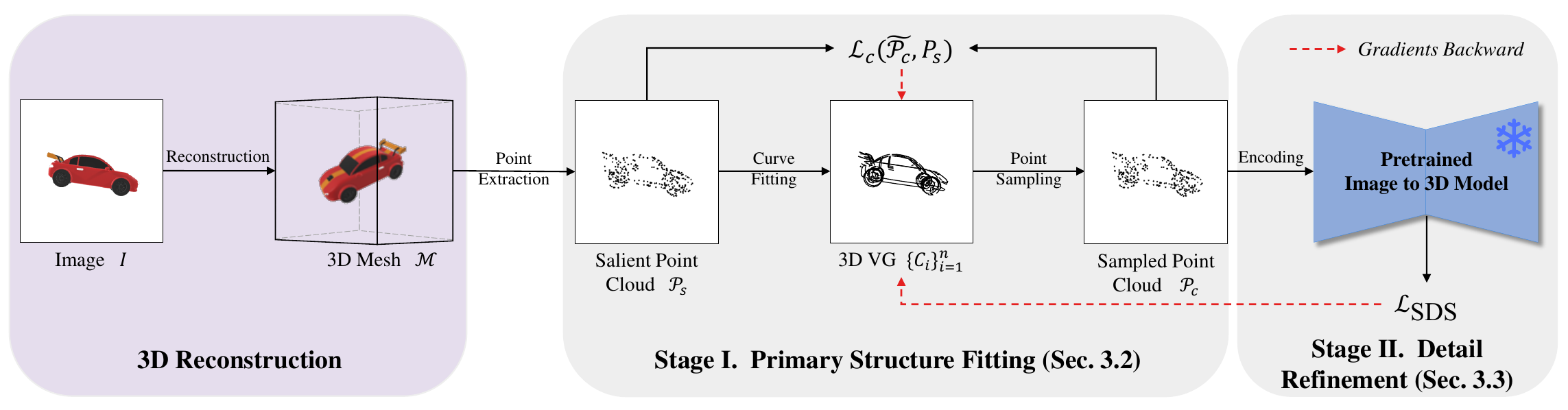}
  \caption{
  The overall architecture of the proposed method, showcasing the initial generation of 3D Vector Graphic (3D VG) from an input image and subsequent detail refinement using a pretrained image-to-3D model.}
  \label{fig:arch}
\end{figure}

Our method workflow is illustrated in Figure~\ref{fig:arch}. It begins by reconstructing a 3D mesh $\mathcal{M}$ from a user-provided image $\bm{I}$ using an image-to-3D  model~\cite{li2025triposg}.
We then apply a two-stage process on the resulting mesh, consisting of Bézier curve fitting followed by detail refinement.
The primary structure fitting stage identifies high-curvature regions in the reconstructed mesh,
converts them into a point cloud, and fits Bézier curves to approximate these structures.
The detail refinement stage re-initializes additional curves in regions overlooked during the first stage and optimizes them using Score Distillation Sampling(SDS) loss~\cite{dreamfusion},
leveraging priors from the diffusion model to guide the optimization and enhance fine-grained details in the resulting 3D vector graphic representation.
This two-stage approach ensures both structural accuracy and high-fidelity detail preservation.

\subsection{ Stage I: Primary Structure Fitting}
\label{stage1}

In this stage, we extract key structural information from the reconstructed mesh $\mathcal{M}$ and use it to fit 3D Bézier curves. The fitting process is further optimized using a specially designed Chamfer Distance loss.

\subsubsection{Salient Point Cloud Extraction}
\label{salient_point}

To identify high-curvature regions on the mesh, we adopt the Sharp Edge Sampling (SES) process from Dora~\cite{chen2024dora} to extract a salient point cloud.
We traverse each edge of the mesh.
For each edge, if it belongs to two adjacent faces,
we compute the angle between the normal vectors of these two faces.
If the angle is below a predefined threshold, we consider this edge as a salient edge.
To address the challenge of extracting salient edges from smooth surfaces (e.g., spheres), we uniformly sample camera parameters on a horizontal plane.
For each sampled viewpoint, we compute the front faces and back faces.
Edges shared by a front face and a back face are identified as silhouette edges.
All such silhouette edges are subsequently incorporated into the salient edge set.
After identifying all salient edges, we sample points along these edges to create a point cloud $\mathcal{P}_{s}$ as ground truth.

\subsubsection{Point Cloud Clustering}

After obtaining the salient point cloud $\mathcal{P}_{s}$, we aggregate these discrete points into clusters suitable for Bézier curve fitting.
Inspired by EdgeGaussians~\cite{edgegaussians}, we perform clustering for edge fitting based on vertex orientations. While EdgeGaussians directly utilize the principal directions of 3D Gaussians as orientation vectors, such directional information is absent in our discrete point cloud $\mathcal{P}_{s}$.
To address this, we introduce an initialization step to estimate orientation vectors for each point in $\mathcal{P}_{s}$. Specifically, for each point $p$, we first identify its $k$ nearest neighbors and then apply Principal Component Analysis (PCA)~\cite{pearson1901liii} to the local neighborhood. The resulting primary eigenvector is used as an approximation of $p$'s orientation vector $\vec{v_p}$.

\begin{figure}[htb]
  \centering
  \includegraphics[width=1\textwidth]{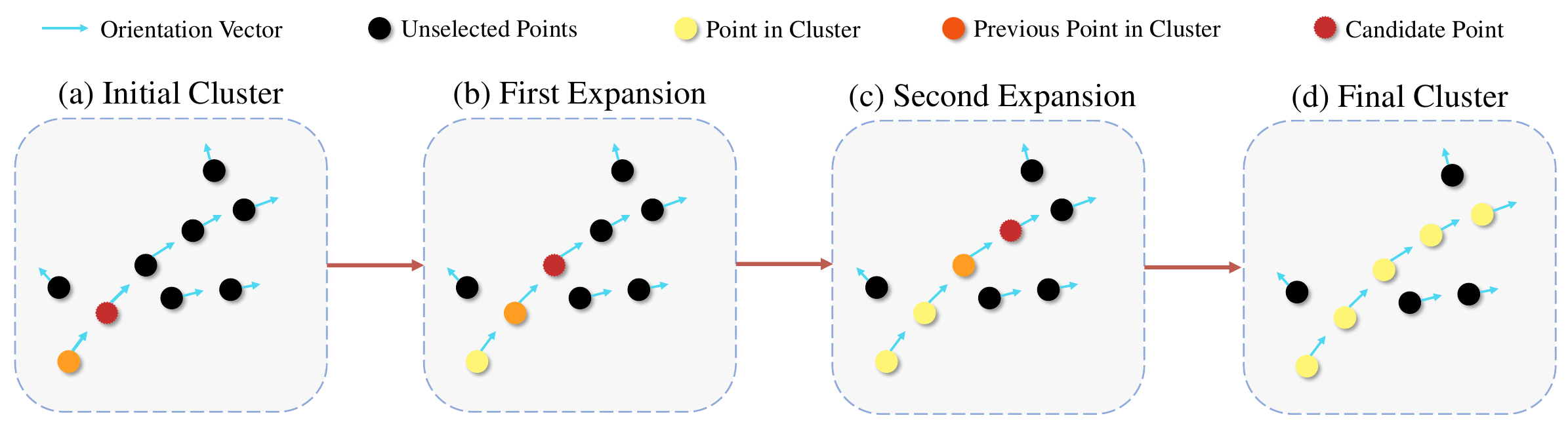}
  \caption{The visualization process of Point Cloud Clustering. Each point is assigned an orientation vector (blue arrows). In (a), the orange point initializes the cluster, and a candidate point (red) is evaluated based on spatial proximity and orientation similarity. In (b), the candidate point meets both criteria and is incorporated into the cluster. This process iterates until the final cluster is formed, as shown in (d).
  }
  \label{fig:cluster}
\end{figure}

Once orientation vectors are assigned to each point, we partition the point cloud into multiple clusters using an iterative expansion process as shown in Fig~\ref{fig:cluster}.
Each cluster is initialized from a randomly selected starting point and grows by progressively incorporating neighboring points that satisfy both spatial proximity and orientation similarity.
Specifically, we use the most recently added point in the cluster, denoted as $p$, to guide the selection of the next candidate. A candidate point $q$ is added to the current cluster if it meets two criteria: (1) it lies within the spatial neighborhood of the cluster, i.e., $dis(p, q) \leq d_{\text{thresh}}$; and (2) the direction of the new edge formed by $p$ and $q$ aligns with the orientation vector of $q$, i.e., $\arccos\left( \left| \frac{\vec{pq}}{\left|\vec{pq}\right|} \cdot \frac{\vec{v_q}}{\left|\vec{v_q}\right|} \right| \right) < \theta_{\text{thresh}}$. The expansion continues until no additional points can be incorporated under these constraints, completing one cluster.

This process repeats across the point cloud to generate a complete set of directional clusters, each representing potential curves for subsequent Bézier fitting. The randomized clustering method ensures comprehensive coverage of all geometric features while avoiding bias toward specific regions. Finally, we filter the small clusters with fewer than $\tau$ points, as they are likely to be noise or outliers.

\subsubsection{Bézier Curves Fitting}
\label{fitting}

After obtaining these clusters,
we attempt to fit them with either straight lines or Bézier curves,
selecting the one with the least error as the fitting result for that cluster.

The fitting process is performed using the Chamfer Distance loss function~\cite{qi2017pointnet} to minimize the distance between the Bézier curves and the salient point cloud $\mathcal{P}_{s}$ obtained in Sec.~\ref{salient_point}.
The fitted Bézier curves are denoted as $\{ C_{i} \}_{i=1}^{n}$.
For our implementation, we use cubic Bézier curves with four control points $P_0, P_1, P_2, P_3 \in \mathbb{R}^{3}$. The parametric equation for a 3D cubic Bézier curve can be written as:

\begin{equation}
    B(t) = (1-t)^3P_0 + 3(1-t)^2tP_1 + 3(1-t)t^2P_2 + t^3P_3, \quad t \in [0,1]
    \label{eq:1}
\end{equation}

To generate a point cloud from a set of Bézier curves $\{ C_i \}_{i=1}^n$,
we uniformly sample $s$ points along each curve by evaluating the parametric function $B(t)$ at $t_j = \frac{j-1}{s-1}$ for $j = 1, \dots, s$.
The resulting point cloud $\mathcal{P}_c$ is defined as:  

\begin{equation}
\mathcal{P}_c = \big\{ B_i(t_j) \mid i \in \{1, \dots, n\}, \, j \in \{1, \dots, s\} \big\}.
\end{equation}

Chamfer Distance loss is computed as follows:

\begin{equation}
	\mathcal{L}_{c}(\tilde{\mathcal{P}_{c}}, P_{s}) = \frac{\lambda}{|\tilde{\mathcal{P}_{c}}|} \sum_{p \in \tilde{\mathcal{P}_{c}}} \min_{q \in \mathcal{P}_{s}} \left\|p - q\right\|^2
	+
	\frac{1}{|\mathcal{P}_{s}|} \sum_{q \in \mathcal{P}_{s}} \min_{p \in \tilde{\mathcal{P}_{c}}} \left\|p - q\right\|^2
\end{equation}

where $\tilde{\mathcal{P}}_{c}$ denotes $\mathcal{P}_{c}$ augmented with Gaussian noise (introduced for data augmentation),
and $\lambda$ is a hyperparameter to balance the two terms.
The generation of point cloud $\mathcal{P}_c$, which relies on the Bézier curve formulation in Eq.~\ref{eq:1}, is differentiable with respect to the curve's control points.
Consequently, the chain rule enables the gradients from $\mathcal{L}_{c}$ to propagate back to these control points, facilitating their iterative optimization.

\subsection{Stage II: Detail Refinement}

Some objects may contain intricate-to-approximate regions that Stage I might miss due to limitations in the salient point cloud extraction or clustering process.
These regions are identified by analyzing the mesh's vertex distribution and locating areas not adequately covered by the point cloud $\mathcal{P}_c$ generated in Stage I.
For such cases, we introduce an additional refinement stage to handle these regions by distilling priors from a pretrained image-to-3D model.



First, the parameters of the initial curves $\{ C_{i} \}_{i=1}^{n}$ from Stage I are frozen.
We randomly initialize new Bézier curves $\{ C'_{i} \}_{i=1}^{n'}$ (with parameters $\theta'$) in regions that are intricate to approximate, thereby complementing the primary structure.
To jointly represent both curve sets, we sample a combined point cloud $\mathcal{P}_{combined} = \mathcal{P}_c \cup \mathcal{P}_{c'}$ and encode it into a latent space $\mathcal{Z}$ using a pretrained VAE encoder from~\cite{li2025triposg}: $z = \mathcal{E}(\mathcal{P}_{combined})$.
Then we refine only the new parameters $\theta'$ via SDS loss~\cite{dreamfusion}, supervised by the input image $\bm{I}$. 


\begin{equation}
    \nabla_{\theta'} \mathcal{L}_{\text{SDS}} = \mathbb{E}_{t,\epsilon} \left[ w(t) \left( \epsilon_\phi(z_t, \bm{I}; t) - \epsilon \right) \frac{\partial z}{\partial \theta'} \right]
\end{equation}

where $z_t$ is the noised latent variable at timestep $t$, $\epsilon_\phi$ is the denoising model conditioned on $\bm{I}$, and $w(t)$ is a weighting function.
This process iteratively adjusts the newly added Bézier curves to fill in missing details from Stage I, ensuring consistency and high fidelity in the final 3D vector representation $\mathcal{S}^{3D} = \{ C_{i} \}_{i=1}^{n} \cup \{ C'_{i} \}_{i=1}^{n'}$.

\subsection{3D Vector Graphics Rendering}

To enable both qualitative visualization and quantitative evaluation, 3D vector graphic $S^{3D}$ should be projected onto a 2D plane.
As proved by 3Doodle~\cite{choi20243doodle}, the perspective projection of a 3D Bézier curve onto a 2D plane yields a 2D rational Bézier curve.
Given a 3D curve $B(t)$ and image plane at $z=f$ (where $z$ is the depth axis and $f$ the focal length), the projection is:

\begin{equation}
    B^{2D}(t) = \left(  B_{x}(t)\frac{f}{B_{z}(t)} \atop B_{y}(t)\frac{f}{B_{z}(t)} \right)
\end{equation}

Consequently, by defining camera parameters in 3D space, we obtain a set of 2D rational Bézier curves corresponding to the camera's viewpoint.
These curves can be rendered using DiffVG~\cite{li2020differentiable} to generate corresponding SVG files, which are subsequently utilized for both quantitative and qualitative analysis.

%% file: exp.tex
\section{Experiment}
\label{exp}

\subsection{Implementation Details}

Our VC3D framework is implemented in PyTorch.
For primary structure fitting stage, we set the distance threshold $d_{\text{thresh}} = 0.05$ and angle threshold $\theta_{\text{thresh}} =  \ang{50}$.
Each Bézier curve is defined by $4$ control points, with $s=64$ sample points per curve for optimization.
For detail refinement stage, we employ TripoSG~\cite{li2025triposg} as our pretrained image-to-3D model, with SDS loss weight set to $2 \times 10^{-4}$. We use the SGD optimizer~\cite{robbins1951stochastic} with a learning rate of $5 \times 10^{-3}$.

All experiments are conducted on a single NVIDIA RTX 4090 GPU.
For each input, our method typically produces fewer than 100 Bézier curves.
Our full method takes about 30 minutes to generate a vector graphic, with 100 optimization steps for Stage I and 200 steps for Stage II.
We collected $40$ images from prior works and online sources as inputs.
All generated 3D vector graphics are rendered into $12$ views using identical camera parameters, upon which the metrics are computed.

\subsection{Experimental Results}

We compare our approach with two state-of-the-art methods in 3D vector graphics generation: Diff3DS~\cite{zhang2024diff3ds}, which designs a depth-aware differentiable rasterizer and leverages 2D diffusion model priors through SDS loss to generate 3D vector graphics from text or images, and 3Doodle~\cite{choi20243doodle}, which employs perceptual losses with multi-view guidance to obtain 3D Bézier curve representations of objects. To comprehensively evaluate the quality and fidelity of the generated 3D vector graphics, we employ CLIPScore~\cite{radford2021learning} to measure semantic alignment between rendered views and input images
Furthermore, we use an aesthetic indicator~\cite{schuhmann2022aesthetic} to quantify the aesthetic value.

\subsubsection{Qualitative Evaluation}
\label{sec:qualitative}

\begin{figure*}[h]
\centering
\includegraphics[width=1\linewidth]{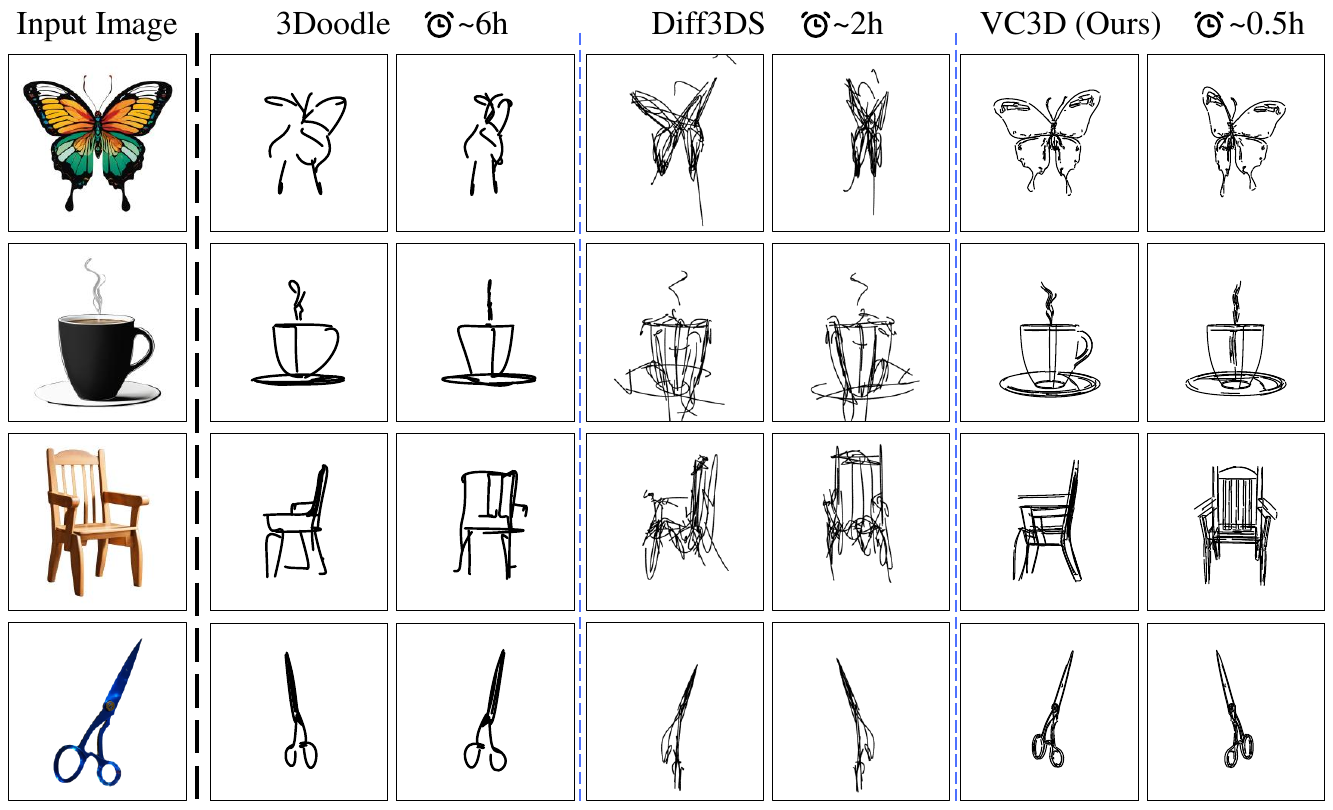}
\caption{
Qualitative comparison of different methods. Diff3DS and VC3D use a single image $\bm{I}$ as input, while 3Doodle uses $120$ rendered images of the mesh reconstruction result $\mathcal{M}$ as input.
}
\label{fig:qualitative}
\end{figure*}

Figure~\ref{fig:qualitative} presents qualitative comparisons between our method and previous work, 3Doodle~\cite{choi20243doodle} and Diff3DS~\cite{zhang2024diff3ds}.
As shown, VC3D produces cleaner, more accurate, and more view-consistent 3D vector graphics.
Previous works struggle to capture fine details in reference images, such as the patterns on butterfly or the handle of the coffee cup. Additionally, their outputs often contain excessive messy lines (e.g., the chair example).

\subsubsection{Quantitative Evaluation}

Table~\ref{tab:quantitative} presents the quantitative analysis results of all methods.
Our method outperforms both previous approaches in CLIPScore and Aesthetic Score metrics.
Our method achieves a cosine similarity of $0.799$, which is higher than the $0.729$ achieved by 3Doodle and the $0.673$ achieved by Diff3DS.
At the same time, we achieved the highest score in Aesthetic Score.
These superior results demonstrate our method's ability to generate semantically and geometrically superior 3D vector graphics.

In addition to the metrics mentioned above, our method demonstrates significant advantages in generation time.
Our method requires only a few SDS loss optimization steps, significantly reducing generation time.
The total runtime for two stages is approximately $0.5$ hours, showing notable improvements compared to 3Doodle (\textasciitilde 6 hours) and Diff3DS (\textasciitilde 2 hours).

\begin{table}[h!]
\centering
\begin{minipage}[t]{0.48\textwidth}
\centering
\caption{Quantitative comparison of VC3D and previous methods on evaluation metrics. The \textbf{bold numbers} represent the best performance.}
\begin{tabular}{>{\centering\arraybackslash}m{2.0cm} >{\centering\arraybackslash}m{1.9cm} >{\centering\arraybackslash}m{1.5cm}}
\toprule
\textbf{Method} & \textbf{CLIPScore $\uparrow$} & \textbf{Aesthetic Score $\uparrow$} \\ \midrule
3Doodle  & 0.729    &  4.122   \\ 
Diff3DS  & 0.673    &  3.769    \\ 
VC3D (Ours)     & \textbf{0.799}  &\textbf{4.352}   \\ 
\bottomrule
\end{tabular}
\vspace{0.2cm}
\label{tab:quantitative}
\end{minipage}
\hspace{0.02\textwidth} 
\begin{minipage}[t]{0.48\textwidth}
\centering
\caption{Ablation study results comparing different variants of our proposed method. The \textbf{bold numbers} represent the best performance.}
\begin{tabular}{>{\centering\arraybackslash}m{2cm} >{\centering\arraybackslash}m{1.9cm} >{\centering\arraybackslash}m{1.5cm}}
\toprule
\textbf{Method} & \textbf{CLIPScore $\uparrow$} & \textbf{Aesthetic Score $\uparrow$} \\ \midrule
Variant 1          & 0.779                        & 4.096                       \\ 
Variant 2         & 0.805                        & 4.167                       \\ 
Full Method             & \textbf{0.818}               & \textbf{4.297}              \\ \bottomrule
\end{tabular}
\vspace{0.2cm}

\label{tab:ablation}
\end{minipage}
\vspace{-0.5cm}
\end{table}

\subsection{Ablation Studies and Analysis}

To demonstrate the respective contributions of the Chamfer Distance loss and SDS loss, we performed ablation experiments.
We selected a subset of $20$ images from the inputs in Section~\ref{sec:qualitative}, where all examples were optimized with SDS loss.
And results were recorded after three distinct stages, corresponding to three variants: (1) \textbf{Variant 1:} This variant refers to the model with the salient point cloud extraction and the point cloud clustering, (2) \textbf{Variant 2:} This variant is the model with the first stage only, \textit{i.e.}, Primary Structure Fitting, and (3) \textbf{Full Method}: This is our proposed method, which comprises two stages.
The results are shown in Table~\ref{tab:ablation}. 
The improvements of the \textbf{Variant 2} against the \textbf{Variant 1} indicate the benefits brought by using CD loss for optimization.
Comparing the performance of \textbf{Variant 2} and \textbf{Full Method}, it is clear that the detail refinement stage can further improve the performance in  
CLIPScore and Aesthetic Score metrics.

\begin{figure}[h]
  \centering
  \vspace{0.2cm}
  \includegraphics[width=1\textwidth]{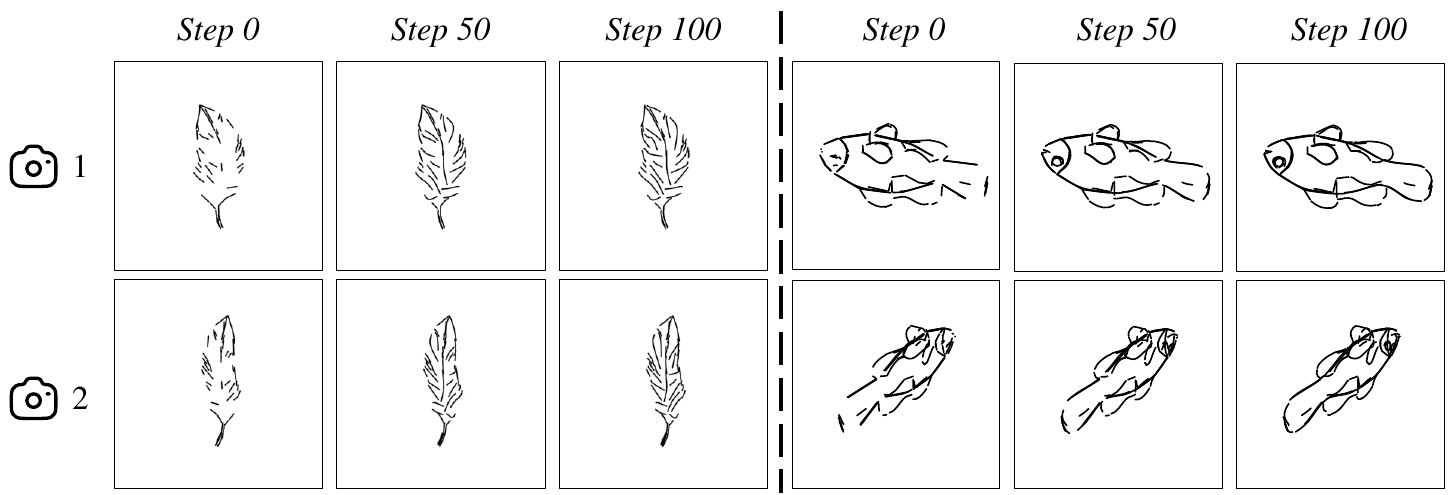}
  \caption{Illustration of the optimization process using Chamfer Distance loss $\mathcal{L}_{c}$.}
  \label{fig:cdloss}
\end{figure}

In Figure~\ref{fig:cdloss}, we show the optimization process of Chamfer Distance loss.
The initially fitted Bézier curves often fail to accurately cover the salient point cloud $\mathcal{P}_s$.
The coherence between curves is also suboptimal.
As optimization progresses, the curves gradually extend to form more complete structures, ultimately achieving both improved coverage of the salient features and enhanced inter-curve coherence while preserving the overall geometric fidelity of the original shape.

\begin{figure}[hbt]
  \centering
  \includegraphics[width=1\textwidth]{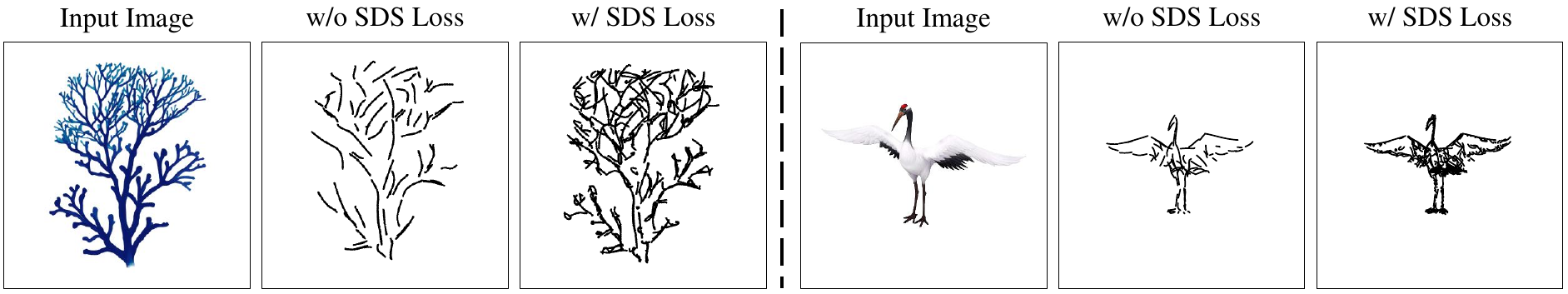}
  \caption{Illustration of the optimization effect of SDS loss $\mathcal{L}_{sds}$. SDS loss effectively recovers missing structural information while enhancing geometric detail representation.}
  \label{fig:sdsloss}
\end{figure}

The visual improvements brought by the SDS loss can be observed in Figure~\ref{fig:sdsloss},
where the refinement stage compensates for previously overlooked details (e.g., terminal branches on corals) and improves the structural coherence of the 3D vector graphics.

These results demonstrate that our two-stage approach effectively balances structural accuracy with visual quality, leading to more compelling and semantically accurate 3D vector graphics.

\begin{figure}[hbt]
  \centering
  \includegraphics[width=1\textwidth]{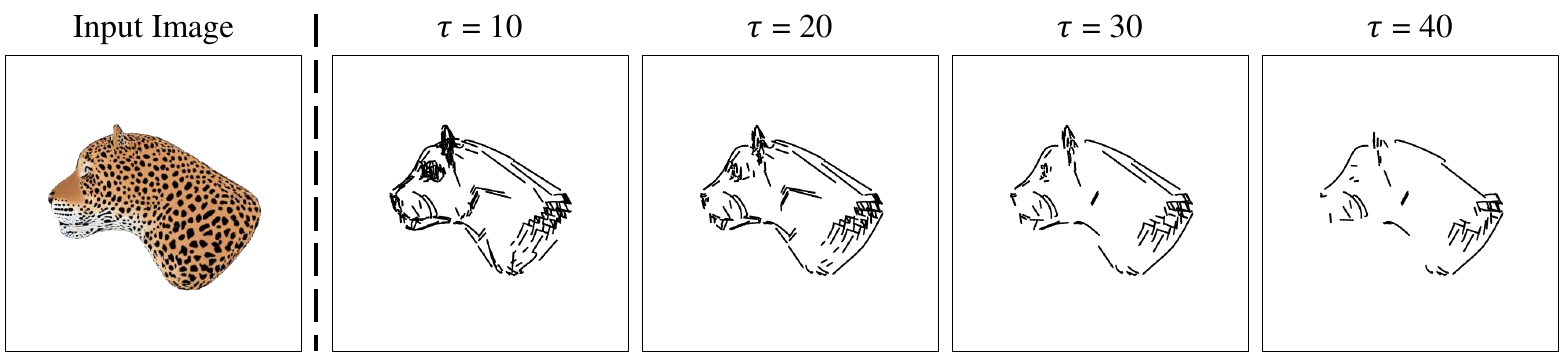}
  \caption{Effect of the filtering threshold in Point Cloud Clustering. The number of Bézier curves gradually decreases as $\tau$ increases, producing a more abstract effect.
  }
  \label{fig:num}
\end{figure}

We also experimented with the number of Bézier curves.
Since the number of Bézier curves is equal to the number of point clusters, we can control the number of curves by adjusting the cluster count, i.e., by changing the filtering threshold $\tau$ in Point Cloud Clustering stage.
As shown in Figure~\ref{fig:num}, when $\tau=10$, point clusters with fewer than 10 points are removed. Increasing the threshold eliminates more clusters, reducing the number of Bézier curves retained and producing an abstract result.  

\section{Limitations and Future Works}
\label{lim}

While VC3D efficiently generates view-consistent 3D vector graphics, it currently lacks occlusion relationships between curves. When rendering to 2D images, all curves share uniform transparency, which may compromise visual fidelity.
Future work could address this by utilizing the corresponding mesh. Since the mesh is available, the position of each Bézier curve relative to the camera parameters can be determined, allowing for the processing of occlusion relationships.

In addition, considering that our method can generate corresponding 3D vector graphics from meshes with minimal time cost, we can build 3D vector graphics datasets based on open-source mesh datasets in the future, providing a research foundation for subsequent work.

%% file: con.tex
\section{Conclusion}
\label{con}

In this paper, we present VC3D, a novel framework for generating view-consistent 3D vector graphics using 3D priors.
Operating directly in 3D space rather than 2D projection planes, our approach effectively addresses view inconsistency issues.
Our two-stage algorithm first identifies salient structures through geometric clustering and Bézier curve fitting,
then refines results using SDS loss with a pretrained image-to-3D model. Experiments demonstrate that VC3D preserves geometric characteristics while maintaining view consistency across viewpoints, with advantages in generation efficiency. This research makes high-quality 3D vector graphics creation more accessible and applicable to virtual reality, shape retrieval, and conceptual design.